\documentclass[11pt,a4paper]{article}
\usepackage[hyperref]{emnlp2020}
\usepackage{times}
\usepackage{latexsym}

\usepackage{array}
\usepackage{boldline}
\usepackage{bm}
\usepackage[utf8]{inputenc}
\usepackage{filecontents}
\usepackage{amsfonts}
\usepackage{multirow}
\usepackage{adjustbox}
\usepackage{amsmath,mathtools}
\usepackage{booktabs}
\usepackage{tikz}
\usepackage{graphicx}
\usepackage{lipsum}

\usepackage[fit]{truncate}
\DeclareGraphicsExtensions{.pdf,.png,.jpg}

\usepackage[ruled,vlined]{algorithm2e}

\usepackage[normalem]{ulem}
\usepackage[T1]{fontenc}
\usepackage[utf8]{inputenc}
\useunder{\uline}{\ul}{}

\usepackage{microtype}

\aclfinalcopy 

\newcolumntype{L}[1]{>{\raggedright\let\newline\\\arraybackslash\hspace{0pt}}m{#1}}
\newcolumntype{C}[1]{>{\centering\let\newline\\\arraybackslash\hspace{0pt}}m{#1}}
\newcolumntype{R}[1]{>{\raggedleft\let\newline\\\arraybackslash\hspace{0pt}}m{#1}}

\newcommand\blfootnote[1]{%
  \begingroup
  \renewcommand\thefootnote{}\footnote{#1}%
  \addtocounter{footnote}{-1}%
  \endgroup
}
\newcommand{\Ours}{adversarial subword regularization}
\newcommand{\ours}{\textsc{AdvSR}}

\newcommand{\base}{\textsc{Base}}
\newcommand{\sr}{\textsc{SR}}

\title{Adversarial Subword Regularization for \\ Robust Neural Machine Translation}

\author{Jungsoo Park \quad Mujeen Sung \quad Jinhyuk Lee$^\dagger$ \quad Jaewoo Kang$^\dagger$\\
Korea University\\
\texttt{\{jungsoo\_park,mujeensung,jinhyuk\_lee,kangj\}@korea.ac.kr}
}

\date{}

\begin{document}
\maketitle
\begin{abstract}
Exposing diverse subword segmentations to neural machine translation (NMT) models often improves the robustness of machine translation as NMT models can experience various subword candidates.
However, the diversification of subword segmentations mostly relies on the pre-trained subword language models from which erroneous segmentations of unseen words are less likely to be sampled.
In this paper, we present \Ours~(\ours) to study whether gradient signals during training can be a substitute criterion for exposing diverse subword segmentations.
We experimentally show that our model-based adversarial samples effectively encourage NMT models to be less sensitive to segmentation errors and improve the performance of NMT models in low-resource and out-domain datasets.
\end{abstract}
\blfootnote{\textsuperscript{$\dagger$}Corresponding authors}

\section{Introduction}
\label{sec:introduction}

Subword segmentation is a method of segmenting an input sentence into a sequence of subword units~\citep{Sennrich2016bpe, Wu2016google, kudo2018sr}.
Segmenting a word to the composition of subwords alleviates the out-of-vocabulary problem while retaining encoded sequence length compactly.
Due to its effectiveness in the open vocabulary set, the method has been applied to many NLP tasks including neural machine translation (NMT) and others~\citep{gehring2017convolutional, vaswani2017attention, devlin2019bert, yang2019xlnet}.

Recently, Byte-Pair-Encoding(BPE)~\citep{Sennrich2016bpe} has become one of the \textit{de facto} subword segmentation methods.
However, as BPE deterministically segments each word into subword units, NMT models with BPE always observe the same segmentation result for each word and often fail to learn diverse morphological features.
In this regard,~\citet{kudo2018sr} proposed subword regularization, a training method that exposes multiple segmentations using a unigram language model.
Starting from machine translation, it has been shown that subword regularization can improve the robustness of NLP models in various tasks~\citep{kim2019subword, provilkov2019bpe, drexler2019subword, muller2019domain}.

\begin{figure}[t!]
\centering
\includegraphics[width=1\linewidth]{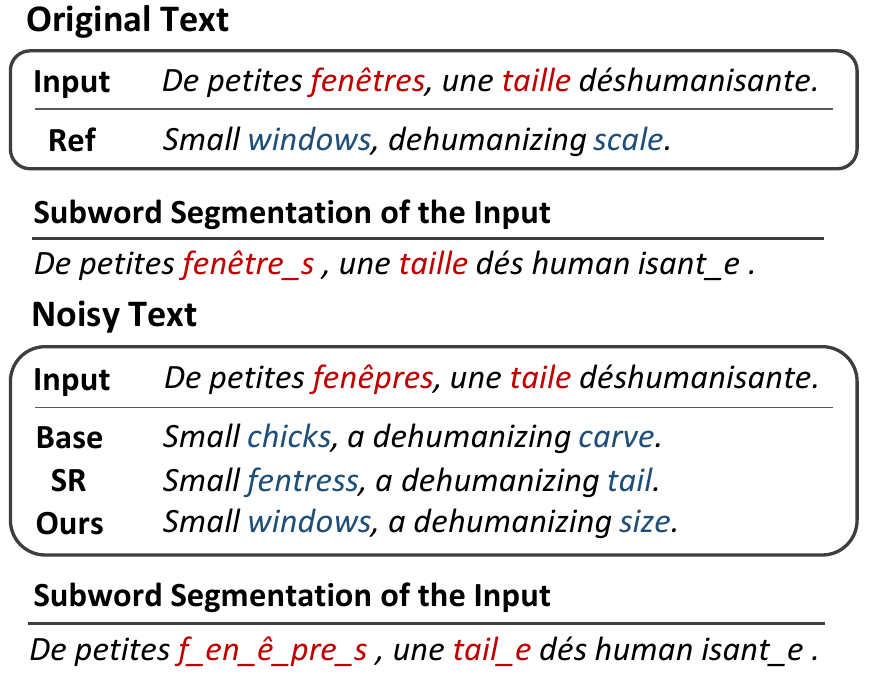}
\caption{NMT models suffer from typos (character drop, character replacement) in the source text due to the unseen subword compositions (`\_' denotes segmentation). On the other hand, \textbf{Ours} correctly decodes them. \textbf{Base}: standard training, \textbf{SR}: subword regularization~\citep{kudo2018sr}}
\label{fig:intro}
\vspace{-.2cm}
\end{figure}

However, subword regularization relies on the unigram language models to sample candidates, where the language models are optimized based on the corpus-level statistics from training data with no regard to the translation task objective.
This causes NMT models to experience a limited set of subword candidates which are frequently observed in the training data. 
Thus, NMT models trained with the subword regularization can fail to inference the meaning of unseen words having unseen segmentations.
This issue can be particularly problematic for low resource languages and noisy text where many morphological variations are not present in the training data. 
The suboptimality issue of the subword segmentation methods has been also raised in many prior works~\citep{kreutzer2018learning, wang2019multilingual, ataman2019latent, salesky2020optimizing}.

 To tackle the problem of unigram language models, we search for a different sampling strategy using \textit{gradient signals} which does not rely on corpus-level statistics and is oriented to the task objective. 
We adopt the adversarial training framework~\citep{Goodfellow2014adv, miyato2016adversarial, ebrahimi2017hotflip, cheng2019robust} to search for a subword segmentation that effectively regularizes the NMT models.
Our proposed method, adversarial subword regularization (\ours), greedily searches for a diverse, yet adversarial subword segmentation which will likely incur the highest translation loss.
Our experiment shows that the NMT models trained with \ours~improve the performance of baseline NMT models up to 3.2 BLEU scores in IWSLT datasets while outperforming the standard subword regularization method.
We also highlight that NMT models trained with the proposed method are highly robust to character-level input noises.\footnote{Our code is available in \url{https://github.com/dmis-lab/AdvSR}}

\section{Background}
\label{sec:background}

\paragraph{Subword Regularization}

Subword regularization~\citep{kudo2018sr} exposes multiple subword candidates during training via on-the-fly data sampling. 
The proposed training method optimizes the parameter set $\theta$ with marginal log-likelihood:

\begin{align}
\label{eq:marginal_likelihood}
\mathcal{L}(\theta)=\sum_{s=1}^{D} \mathbb{E}_{\genfrac{}{}{0pt}{1}{\mathbf{x} \sim P_{seg}(\mathbf{x} | X^{(s)})}{\mathbf{y} \sim P_{seg}(\mathbf{y} | Y^{(s)})}} [\log P(\mathbf{y} |\mathbf{x} ; \theta)]
\end{align}

\noindent where $\mathbf{x}=(x_1, \dots, x_M)$ and $\mathbf{y}=(y_1, \dots, y_N)$ are sampled segmentations (in a subword unit) from a source sentence $X$ and a target sentence $Y$ through the unigram language model~(subword-level) $P_{seg}(\cdot)$ and $D$ denotes the number of samples. 
Generally, a single sample per epoch is used during training to approximate Eq~\ref{eq:marginal_likelihood}.

The probability of a tokenized output is obtained by the product of each subword's occurrence probability where subword occurrence probabilities are attained through the Bayesian EM algorithm~\citep{dempster1977maximum, liang2007infinite, liang2009online}.
Segmentation output with maximum probability is acquired by using Viterbi algorithm~\citep{viterbi1967error}.

\label{sec:Adv_NLP}
\paragraph{Adversarial Regularization in NLP} Adversarial samples are constructed by corrupting the original input with a small perturbation which distorts the model output. \citet{miyato2016adversarial}~adopted the adversarial training framework to the task of text classification where input embeddings are perturbed with adversarial noise $\hat{r}$:

\begin{align}
\label{eq:embedding_perturbation}
e_{i}^{\prime}&=Ex_{i}+\hat{r}_{i} \\
\label{eq:adversarial_objective}
\text {where, }\hat{r}&=\underset{r,\|r\| \leq \epsilon}{\operatorname{argmax}}\{\ell(X, r, Y;\theta)\}
\end{align}

\noindent $E$ is an embedding matrix, $e_{i}^{\prime}$ is an perturbed embedding vector, and $\ell(\cdot)$ is loss function obtained with the input embeddings perturbed with noise $r$.
Note that \citet{miyato2016adversarial} use a word for the unit of $x_i$ unlike our definition.
As it is computationally expensive to exactly estimate $\hat{r}$ in Eq~\ref{eq:adversarial_objective},~\citet{miyato2016adversarial} resort to the linear approximation method~\citep{Goodfellow2014adv}, where $\hat{r_i}$ is approximated as follows:

\begin{align}
\label{eq:linear_approximation}
\hat{r}_{i}=\epsilon \frac{g_{i}}{\|g\|_{2}}, \quad g_{i}=\nabla_{e_{i}} \ell(X, Y;\theta)
\end{align}

\noindent $\epsilon$ indicates the degree of perturbation and $g_i$ denotes a gradient of the loss function with respect to a word vector. 
Moreover,~\citet{ebrahimi2017hotflip} extended adversarial training framework to directly perturb discrete input space, i.e. character, through the first-order approximation by the use of gradient signals.

\section{Approach}
\label{sec:Approach}
Relying on the subword language models might bias NMT models to frequent segmentations, hence hinders the NMT model in understanding diverse segmentations. 
This may harm the translation quality of the NMT models when diverse morphological variations occur.

However, simply exposing diverse segmentations uniformly leads to a decrease in performance~\citep{kudo2018sr}. 
In this regard, we utilize gradient signals for exposing diverse, yet adversarial subword segmentation inputs for effectively regularizing NMT models.
\citet{kreutzer2018learning} proposed to jointly learn to segment and translate by using hierarchical RNN~\citep{graves2016adaptive}, but the method is not model-agnostic and slow due to the increased sequence length of character-level inputs.
On the other hand, our method is model-agnostic and operates on the word-level. Our method seeks adversarial segmentations on-the-fly, thus the model chooses the subword candidates that are vulnerable to itself according to the state of the model at each training step.


\subsection{Problem Definition}
\label{sec:Adv_SR}

Our method generates a sequence of subwords by greedily replacing the word's original segmentation to that of adversarial ones estimated by gradients.
Given a source sentence $X$ and a target sentence $Y$, we want to find the sequence of subwords $\mathbf{\hat{x}}$ and $\mathbf{\hat{y}}$ which incurs the highest loss:

\begin{align}
\label{eq:adv_objective}
\mathbf{\hat{x}}, \mathbf{\hat{y}} &=\underset{{\genfrac{}{}{0pt}{1}{\mathbf{x}\in \Omega(X)}{\mathbf{y}\in \Omega(Y)}}}{\operatorname{argmax}}\{\ell(\mathbf{x}, \mathbf{y} ;\theta)\}
\end{align}

\noindent $\Omega(X)$ and $\Omega(Y)$ denote all the subword segmentation candidates of $X$ and $Y$ and $\ell(\cdot)$ denotes loss function. 

Our method operates on a word unit split by whitespaces, each of which consists of variable length subwords.
We first define a sequence of words in $X$ as $\mathbf{w} = (w_1,\dots,w_{M'})$ where $M'$ denotes the length of the word-level sequence.
Then, we can segment $w_j$ as $\mathbf{s}_j = (s_1^j,\dots,s_K^j)$ which are $K$ subword units of the $j$-th word (note that now we can represent input $X$ as as a sequence of $\mathbf{s}_j$ as $\mathbf{s} = (\mathbf{s}_1,\dots,\mathbf{s}_{M'})$).
For example, as for the $j$-th word \textit{"lovely"}, its tokenized output \textit{"love"} and \textit{"ly"} will be $s_1^{j}$ and $s_2^{j}$ respectively. 
Then, we define the embedding and the gradient of the word segmentation as the aggregation of $K$ subwords consisting it:

\begin{align}
\label{eq:define_embedding}
e(\mathbf{s}_{j})  &= f([e(s_{1}^{j}), \dots, e(s_{K}^{j})]) \in{\mathbb{R}^{d}}  \\
\label{eq:define_gradient}
g_{\mathbf{s}_{j}} &= f([g_{s_{1}^{j}}, \dots, g_{s_{K}^{j}}]) \in{\mathbb{R}^{d}} \\
\label{eq:gradient_embedding}
\text{where}~~g_{s_k^j} &=\nabla_{e\left(s_k^j\right)}\ell(\mathbf{x},\mathbf{y} ;\theta) \in{\mathbb{R}^{d}}
\end{align}

\noindent where $e$ denotes the embedding lookup operation, $d$ denotes the hidden dimension of embeddings.
We simply use the element-wise average operation for $f$.
Therefore if the segmentation of the word changes, the corresponding embedding and gradient vector will change accordingly.

\SetAlFnt{\small}
\setlength{\textfloatsep}{10pt}
\begin{algorithm}[t!]
\SetKwInOut{Input}{input}
\SetKwInOut{Output}{output}
\SetKwProg{Fn}{Function}{:}{}
\Input{$\text{input sentence}~X,~\text{probability}~R$}
\Output{$\text{adversarial subword sequence}~\mathbf{\hat{x}}$}
\Fn{AdvSR(X, $R$)}{
$\mathbf{\hat{x}} \gets [~]~\text{// initialize empty list}$ \\
$\mathbf{\tilde{x}} \gets \underset{\mathbf{x} \in \Omega(X)}{\operatorname{argmax}}~P_{seg}{(\mathbf{x}|X)}$ \\ 
$\mathbf{\tilde{s}} \gets \text{group}(\mathbf{\tilde{x}})~\text{// group subwords as word-level}$  \\
\For{$j\gets1$ \KwTo $M'$}{
    $r \gets \text{uniform}(0,1)$ \\
    \If{$r < R$}{
    $\text{// compute Eq}~\ref{eq:define_gradient}$. \\
    $g_{\Tilde{\mathbf{s}}_{j}} \gets f([g_{\Tilde{s}_{1}^{j}}, \dots, g_{\Tilde{s}_{K}^{j}}])~$\\
    $\text{// compute Eq}~\ref{eq:gradient_objective}.$\\
    $\hat{\mathbf{s}}_j \,\, \gets \underset{\mathbf{s}_j \in \Omega(w_j)}{\operatorname{argmax}} ~ g_{\Tilde{\mathbf{s}}_j}^T\cdot [e({\mathbf{s}_j})-e(\Tilde{\mathbf{s}}_j)]~$\\
    }
    \Else{
    $\hat{\mathbf{s}}_j \,\,\gets \Tilde{\mathbf{s}}_j$ 
    }
    $\mathbf{\hat{x}} \gets \mathbf{\hat{x}} + \hat{\mathbf{s}}_j~\text{// append}$
 }
}
\Return {$\mathbf{\hat{x}}$}
\caption{\textit{AdvSR} function}
\label{alg:pseudo_code}
\end{algorithm}

\subsection{Adversarial Subword Regularization}

As it is intractable to find the most adversarial sequence of subwords given combinatorially large space, we approximately search for word-wise adversarial segmentation candidates.
We seek for the adversarial segmented result of a $j$-th word, i.e. $w_{j}$, from the sentence $X$ by following criteria which was originally proposed by~\citet{ebrahimi2017hotflip} and applied to many other NLP tasks~\citep{cheng2019robust, wallace2019universal, michel2019evaluation}.
More formally, we seek an adversarial segmentation $\hat{\mathbf{s}}_{j}$ of the $j$-th word $w_j$ as

\begin{align}
\label{eq:gradient_objective}
\hat{\mathbf{s}}_{j} &=\underset{\mathbf{s}_j \in \Omega(w_j)}{\operatorname{argmax}}~g_{\Tilde{\mathbf{s}}_j}^T \cdot [e({\mathbf{s}_j})-e(\Tilde{\mathbf{s}}_j)]
\end{align}

\noindent where $\mathbf{s}_j$ represents one of the tokenized output among the possible candidates $\Omega(w_j)$ which are obtained by SentencePiece tokenizer~\citep{kudo2018sentencepiece}.
$\Tilde{\mathbf{s}}_j$ denotes an original deterministic segmentation of $j$-th word.
Note that for computing $g_{\Tilde{\mathbf{s}}_j}$, we use $\ell(\tilde{\mathbf{x}}, \tilde{\mathbf{y}})$ which is from the original deterministic segmentation results.
We applied L2 normalization to the gradient vectors and embedding vectors.

We uniformly select words in the sentence with a probability $R$ and replace them into adversarial subword composition according to the Eq~\ref{eq:gradient_objective}.
We perturb both the source and the target sequences. We summarize our method in Algorithm~\ref{alg:pseudo_code}.
The existing adversarial training methods in the NLP domain generally train the model with both the original samples and the adversarial samples~\citep{miyato2016adversarial, ebrahimi2017hotflip,cheng2019robust, sato2019adv}.
However, we train the model with only the adversarial samples for the sake of fair comparison with the baselines.
More details are described in Appendix~\ref{append:training_details}.

\begin{table}[t!]
\centering
\resizebox{0.42\textwidth}{!}{
\begin{tabular}{cccccc}
\toprule
\textbf{Dataset}  & \textbf{Lang Pair} & \textbf{\shortstack{Number of sentences \\ (train/valid/test)}}  \\ \midrule
\textbf{IWSLT17} & FR $\leftrightarrow$ EN & 232k /  890 / 1210  \\ 
& AR $\leftrightarrow$ EN & 231k /  888 / 1205   \\
\textbf{IWSLT15} & CS $\leftrightarrow$ EN & 105k  / 1385 / 1327 \\ 
                 & VI $\leftrightarrow$ EN & 133k  / 1553 / 1268   \\
\textbf{IWSLT13} & TR $\leftrightarrow$ EN &  132k / 887 /  1568 \\ 
& PL $\leftrightarrow$ EN & 144k  / 767 / 1564  \\   
\textbf{MTNT1.1} & FR $\rightarrow$ EN &  19k / 886 / 1022 (1233)  \\
                 & EN $\rightarrow$ FR &  35k / 852 / 1020 (1401)  \\ \bottomrule
\end{tabular}}
\caption{Data statistics. The number in the parentheses denotes the number of sentences in the MTNT2019 test set which was provided by the WMT Robustness Shared Task~\citep{Xian2019wmt}} 
\label{append:data_statistics}
\vspace{-.15cm}
\end{table}

\begin{table}[t!]
\centering
\resizebox{0.42\textwidth}{!}{
\def\arraystretch{0.8}
\begin{tabular}{C{1.8cm} | C{1.6cm} C{1.3cm} C{1.3cm} C{1.3cm}}
\toprule
\textbf{Lang Pair} &  \textbf{\base} & \textbf{\sr} & \textbf{\ours} \\ \midrule
\multicolumn{4}{c}{\textbf{IWSLT17}}   \\ \midrule
FR $\rightarrow$ EN& 37.9 & \textbf{38.1} & \textbf{38.5} \\ 
EN $\rightarrow$ FR& 38.8 & 39.1 & \textbf{39.8} \\ 
AR $\rightarrow$ EN& 31.7 & \textbf{32.3} & \textbf{32.6} \\
EN $\rightarrow$ AR& \textbf{14.4} & 14.3 & \textbf{14.9} \\    \midrule
\multicolumn{4}{c}{\textbf{IWSLT15}}   \\ \midrule
CS $\rightarrow$ EN& 28.9 & 30.5 & \textbf{32.1} \\ 
EN $\rightarrow$ CS& 20.4 & 21.7 & \textbf{23.0} \\ 
VI $\rightarrow$ EN& 28.1 & 28.4 & \textbf{29.3} \\
EN $\rightarrow$ VI& 30.9 & 31.7 & \textbf{32.4} \\ \midrule
\multicolumn{4}{c}{\textbf{IWSLT13}}   \\ \midrule
PL $\rightarrow$ EN& 19.1 & 19.7 & \textbf{20.6} \\ 
EN $\rightarrow$ PL& 13.5 & 14.1 & \textbf{15.1} \\
TR $\rightarrow$ EN& 21.3 & 22.6 & \textbf{24.0} \\ 
EN $\rightarrow$ TR& 12.6 & \textbf{14.4} & \textbf{14.6} \\
 \bottomrule
\end{tabular}}
\caption{BLEU scores on the main results. Bold indicates the best score
and all scores whose difference from the best is not statistically significant computed via bootstrapping~\citep{koehn2004statistical} ($p$-value $<$ 0.05).}
\label{tab:main_result}
\end{table}

\section{Experimental Setup}
\label{sec:setup}

\subsection{Datasets and Implementation Details}
\label{sec:dataset}

We conduct experiments on a low-resource multilingual dataset, IWSLT\footnote{\url{ http://iwslt.org/}}, where unseen morphological variations outside the training dataset can occur frequently. 
We also test NMT models on MTNT~\citep{Michel2018mtnt}, a testbed for evaluating the NMT systems on the noisy text. 
We used the English-French language pair. 
Moreover, for evaluating the robustness to the typos, we generate the synthetic test data with character-level noises using the IWSLT dataset. 

For all experiments, we use Transformer-Base~\citep{vaswani2017attention} as a backbone model (L=6, H=512) and follow the same regularization and optimization procedures. We train our models with a joined dictionary of the size 16k.
Our implementation is based on Fairseq~\citep{ott2019fairseq}.
Further details on the experimental setup are described in Appendix~\ref{append:experimental_details}.

\subsection{Evaluation}
\label{sec:evaluation}

For inference, we use a beam search with a beam size of 4. 
For the evaluation, we used the checkpoint which performed the best in the validation dataset. We evaluated the translation quality through BLEU~\citep{papineni2002bleu} computed by SacreBleu~\citep{post2018call}.
Our baselines are NMT models trained with deterministic segmentations (\base) and models trained with the subword regularization method (\sr)~\citep{kudo2018sr}.
We set the hyperparameters of subword regularization equivalent to those of~\citet{kudo2018sr}.

\section{Experiments}
\label{sec:experiments}

\subsection{Results on Low-Resource Dataset}
\label{sec:main_result}

Table \ref{tab:main_result} shows the main results on IWSLT datasets.
Our method significantly outperforms both the \base~and the \sr. This shows that leveraging translation loss to expose various segmentations is more effective than constraining the NMT models to observe limited sets of segmentations.
Specifically, \ours~improves 1.6 BLEU over \sr~and 3.2 BLEU over \base~in the Czech to English dataset.
We assume that the large gains are due to the morphological richness of Czech.
The performance improvement over the baselines can also be explained by the robustness to unseen lexical variations, which are shown in Appendix \ref{append:translation_outputs}. 

\begin{table}[t!]
\centering
\resizebox{0.42\textwidth}{!}{
\def\arraystretch{0.8}
\begin{tabular}{C{1.8cm} | C{1.6cm} C{1.3cm} C{1.3cm}}
\toprule
\textbf{Dataset}  & \textbf{\base} & \textbf{\sr} & \textbf{\ours} \\ \midrule
\multicolumn{4}{c}{\textbf{MTNT2018}}   \\ \midrule
FR $\rightarrow$ EN & 25.7 & \textbf{27.6} & \textbf{27.2} \\ 
EN $\rightarrow$ FR & 26.7 & 27.5 & \textbf{28.2} \\  \midrule
\multicolumn{4}{c}{\textbf{MTNT2018 + FT}}   \\ \midrule
FR $\rightarrow$ EN &  36.5 & 37.9  &  \textbf{38.8} \\
EN $\rightarrow$ FR & 33.2 & 34.4 & \textbf{35.3} \\ \midrule
\multicolumn{4}{c}{\textbf{MTNT2019}}   \\ \midrule
FR $\rightarrow$ EN & 27.6 & 29.3 & \textbf{30.2} \\
EN $\rightarrow$ FR & 22.8 & \textbf{23.8} & \textbf{24.1} \\  \midrule
\multicolumn{4}{c}{\textbf{MTNT2019 + FT}}   \\ \midrule
FR $\rightarrow$ EN &  36.2 & \textbf{38.1} &  \textbf{38.6} \\
EN $\rightarrow$ FR &  27.6 & 28.2 &  \textbf{28.9} \\  \bottomrule
\end{tabular}}
\caption{BLEU scores on the MTNT~\citep{Michel2018mtnt} dataset.  \textbf{FT} denotes finetuning.}
\label{tab:mtnt}
\vspace{-.15cm}
\end{table}

\subsection{Results on Out-Domain Dataset}
\label{sec:mtnt}

Table \ref{tab:mtnt} shows the results on the MTNT dataset where we utilized the NMT models trained from Section \ref{sec:main_result}. We also experiment with the domain adaptive fine-tuning with the MTNT dataset (denoted as \textbf{+ FT}).

Generally, exposing multiple subword candidates to the NMT models shows superior performance in domain adaptation, which matches the finding from~\citet{muller2019domain}. Above all, NMT models trained with our proposed method outperforms \base~up to 2.3 and \sr~up to 0.9 BLEU scores. 

\begin{table}[t!]
\centering
\resizebox{0.42\textwidth}{!}{
\def\arraystretch{0.8}
\begin{tabular}{C{1.3cm}|C{0.75cm}C{0.75cm}C{0.75cm}C{0.75cm}C{0.75cm}}
\toprule
\textbf{Method} & \textbf{0.1} & \textbf{0.2} & \textbf{0.3} & \textbf{0.4} & \textbf{0.5}\\
\midrule
\multicolumn{6}{c}{\textbf{FR $\rightarrow$ EN}}   \\ \midrule
\base & 30.7 & 25.6 & 20.3 & 16.2 & 11.4  \\
\sr   & 33.2 & 28.5 & 23.3 & 18.7 & 14.7  \\
\ours  & \textbf{34.8} & \textbf{31.1} & \textbf{28.7} & \textbf{25.0} & \textbf{21.8}  \\ \midrule
\multicolumn{6}{c}{\textbf{EN $\rightarrow$ FR}}   \\ \midrule
\base & 31.1 & 24.2 & 18.6 & 14.6 & 10.6  \\
\sr   & 34.2 & 27.8 & 23.9 & 18.9 & 14.4  \\
\ours  & \textbf{35.1} & \textbf{30.3} & \textbf{26.4} & \textbf{23.0} & \textbf{19.1} \\ \bottomrule
\end{tabular}}
\caption{BLEU scores on the synthetic dataset of typos. The column lists results for different noise fractions.}
\label{tab:synthetic_result}
\end{table}

\subsection{Results on Synthetic Dataset}
\label{sec:noisy_data}

Additionally, we conduct an experiment to see the changes in translation quality according to different noise ratios. 
Using IWSLT17~(FR $\leftrightarrow$ EN), we synthetically generated 3 types of noise, \textbf{1. character drop}, \textbf{2. character replacement}, \textbf{3. character insertion} and perturbed each word with the given noise probability. Table \ref{tab:synthetic_result} shows that as the noise fraction increases, our method proves its robustness compared to the baseline models improving \base~up to 10.4 and \sr~up to 7.1 BLEU scores. 

\section{Related Work}
\label{sec:related_work}

Subword segmentation has been widely used as a standard in the NMT community since the Byte-Pair-Encoding~\citep{Sennrich2016bpe} was proposed. 
\citet{kudo2018sr} introduced the training method of subword regularization.
Most recently, the BPE-dropout~\citep{provilkov2019bpe} was introduced which modifies the original BPE's encoding process to enable stochastic segmentation. 
Our work shares the motivation of exposing diverse subword candidates to the NMT models with previous works but differs in that our method uses gradient signals. 
Other segmentation methods include wordpiece~\citep{Schuster2012japkor} and variable length encoding schme~\citep{Chitnis2015var}.
Also, there is another line of research that utilizes character-level segmentation~\citep{Luong2016char, Lee2017char, Cherry2017char}.
 
Other works explored generating synthetic or natural noise for regularizing NMT models~\citep{Belinkov2018noise, Sperber2019noise, Karpukhin2019noise}. \citet{Michel2018mtnt} introduced a dataset scraped from Reddit for testing the NMT systems on the noisy text. 
Recently, a shared task on building the robust NMT models was held~\citep{Xian2019wmt, Berand2019naver}.

Our method extends the adversarial training framework, which was initially developed in the vision domain~\citep{Goodfellow2014adv} and has begun to be adopted in the NLP domain recently~\citep{jia2017adversarial, Belinkov2018noise, samanta2017towards, miyato2016adversarial, michel2019evaluation, sato2019adv, wang2019robust, cheng2019robust}.  \citet{miyato2016adversarial} adopted the adversarial training framework on text classification by perturbing embedding space with continuous adversarial noise. 
\citet{cheng2019robust} introduced an adversarial training framework by discrete word replacements where candidates were generated from the language model. However, our method does not replace the word but replaces its subword composition.
 
\section{Conclusions}

In this study, we propose adversarial subword regularization which samples subword segmentations that maximize the translation loss.
Segmentations from the subword language model might bias NMT models to frequent segmentations in the training set.
On the other hand, our method regularizes the NMT models to be invariant to unseen segmentations. Experimental results on low resource and out-domain datasets demonstrate the effectiveness of our method.

\section*{Acknowledgement}
This research was supported by the National Research Foundation of Korea (NRF-2020R1A2C3010638, NRF-2016M3A9A7916996) and Korea Health Technology R\&D Project through the Korea Health Industry Development Institute (KHIDI), funded by the Ministry of Health \& Welfare, Republic of Korea (grant number: HR20C0021).


\bibliography{anthology,emnlp2020}
\bibliographystyle{acl_natbib}

\clearpage
\appendix
\numberwithin{table}{section}
\setcounter{page}{1}


\section{Implementation Details}
\label{append:implementation_details}

\subsection{Details of Training}
\label{append:training_details}

During training, we set $R=\{0.25, 0.33\}$ based on the validation performance.
The words which are not perturbed according to adversarial criterion are deterministically segmented by the SentencePiece.
Note that no other hyper-parameters are tuned. 

We use SentencePiece~\citep{kudo2018sentencepiece} toolkit for acquiring a pre-defined number of subword candidates where we generated up to 9 segmentation candidates per word. 
We use the same SentencePiece tokenizer for training \sr~and for generating segmentation candidates from \ours.  

While training, translation pairs were batched together by their sequence lengths. For all the experiments, the values of batch sizes (number of source tokens) is set to 4096.
All our experiments were conducted with a single GPU~(TitanXP or Tesla P40) and accumulated gradients for 8 training steps. 
Note that the number of parameters of the model (i.e. Transformer Base) is the same for the baselines and our method.

\subsection{Details of Experimental Settings}
\label{append:experimental_details}

Multilingual dataset IWSLT can be downloaded from \url{https://wit3.fbk.eu/} and the MTNT dataset can be downloaded from \url{https://www.cs.cmu.edu/~pmichel1/mtnt/}.
We use the training and validation dataset of MTNT 2018 version for finetuning our model in Section \ref{sec:mtnt}. 
To be specific, we finetune each NMT model in Section \ref{sec:main_result} for 30 epochs. 
We utilized the checkpoint which performed best in the MTNT validation dataset.

Also, for experimenting the \sr, we set the hyper-parameters \textit{alpha} and \textit{l} as 0.1 and 64, respectively which is equivalent to that of original paper. 
Byte Pair Encoding~\citep{Sennrich2016bpe} is not used as the baseline model since the performance is almost the same as that of \base. 
\citet{kudo2018sr} also report scores using n-best decoding, which averages scores from n-best segmentation results. 
However, n-best decoding is n-times time consuming compared to the standard decoding method. 
Therefore we only use 1-best decoding which is the standard decoding framework for evaluating the translation quality.
Our BLEU scores are calculated through SacreBLEU where our signature is as follows:\newline~{\tt BLEU+case.lc+lang.[src-lang]\newline-[dst-lang]+numrefs.1+smooth.exp\newline+tok.13a+version.1.4.2}

\clearpage
\onecolumn

\section{Sampled Translation Outputs}
\label{append:translation_outputs}

\begin{table}[h!]
\centering
\resizebox{0.9\textwidth}{!}{
\begin{tabular}{C{1.5cm} | C{5cm} C{5cm} C{4cm}||C{4cm}C{4cm} C{4cm}C{4cm}} 
\multicolumn{1}{c}{} & \multicolumn{3}{l}{PL$\rightarrow$EN} & \multicolumn{2}{l}{CS$\rightarrow$EN} & \multicolumn{2}{l}{FR$\rightarrow$EN} \\ \hlineB{3}
Input & \multicolumn{3}{l|}{Chodź, \textbf{zatańcz} ze mną.} & \multicolumn{2}{l|}{My \textbf{aktivujeme} komunitu.} & \multicolumn{2}{l}{\textbf{Profitez} de votre soirée.}\\ \hline
Seg. & \multicolumn{3}{l|}{Chodź , \textbf{za\_ta\_ń\_cz} ze mną} & \multicolumn{2}{l|}{My \textbf{aktiv\_ujeme} komunitu .} & \multicolumn{2}{l}{\textbf{Pro\_fi\_t\_ez} de votre soirée .} \\ \hline
\textsc{Ref.} & \multicolumn{3}{l|}{Come, \textbf{dance} with me.} & \multicolumn{2}{l|}{We \textbf{activate} the community.} & \multicolumn{2}{l}{\textbf{Enjoy} your night.}\\ 
\base & \multicolumn{3}{l|}{Come with me}  & \multicolumn{2}{l|}{We \textbf{act} the community.} & \multicolumn{2}{l}{\textbf{Get out of} your night.}\\ 
\sr & \multicolumn{3}{l|}{Come on. \textbf{Stay} with me.}    & \multicolumn{2}{l|}{We \textbf{act} a community.} & \multicolumn{2}{l}{\textbf{Protect} your evening.}\\ 
\ours & \multicolumn{3}{l|}{Come, \textbf{dance} with me.} & \multicolumn{2}{l|}{We \textbf{activate} the community.} & \multicolumn{2}{l}{\textbf{Enjoy} your evening.}\\ \hlineB{3}

\end{tabular}}
\caption{Excerpt from the translation results of the NMT models trained with different training methods. Presented samples demonstrate how our method infers the meaning of rarely appearing words' variations. Despite its low frequency of appearance, the NMT model trained with our method infers the meaning of the observed word's morphosyntactic variation. This can be explained by the fact that our method encourages the NMT model to be segmentation invariant, and is better at inferring the meaning from unseen subword composition.}
\end{table}

\end{document}